\documentclass{article}
\usepackage{spconf,amsmath,graphicx}
\usepackage[]{algorithm2e}
\usepackage{color}
\usepackage{lineno,hyperref}
\usepackage{verbatim}
\usepackage{booktabs}

\usepackage{multirow}
\renewcommand{\vec}[1]{\mathbf{#1}}

\newcommand{\xx}{\vec x}
\newcommand{\vold}[1]{$#1\!\times\!#1\!\times\!#1$}

\title{Isointense infant brain segmentation with a hyper-dense connected convolutional neural network}
 \name{Jose Dolz$^{\star}$ \qquad Ismail Ben Ayed$^{\star}$ \qquad Jing Yuan$^{\dagger}$ \qquad Christian Desrosiers$^{\star}$}

 \address{$^{\star}$ LIVIA Laboratory, \'Ecole de technologie sup\'erieure (ETS), Montreal, QC, Canada\\
 $^{\dagger}$ Xidian University, School of Mathematics and Statistics, Xi'an, China}

\begin{document}
%
\maketitle

\begin{abstract}
Neonatal brain segmentation in magnetic resonance (MR) is a challenging problem due to poor image quality and low contrast between white and gray matter regions. Most existing approaches for this problem are based on multi-atlas label fusion strategies, which are time-consuming and sensitive to registration errors. As alternative to these methods, we propose a hyper-densely connected 3D convolutional neural network that employs MR-T1 and T2 images as input, which are processed independently in two separated paths. An important difference with previous densely connected networks is the use of direct connections between layers from the same and different paths. Adopting such dense connectivity helps the learning process by including deep supervision and improving gradient flow. We evaluated our approach on data from the MICCAI Grand Challenge on 6-month infant Brain MRI Segmentation (iSEG), obtaining very competitive results. Among 21 teams, our approach ranked first or second in most metrics, translating into a state-of-the-art performance.
\end{abstract}
\section{Introduction}
\label{sec:intro}

The precise segmentation of infant brain images into white matter (WM), gray matter (GM) and cerebrospinal fluid (CSF) during the first year of life is of great importance in the study of early brain development. Recognizing particular brain abnormalities shortly after birth might allow to predict neuro-developmental disorders. To that end, magnetic resonance imaging (MRI) is the preferred modality for imaging neonatal brain because it is safe, non-invasive and provides cross-sectional views of the brain in multiple contrasts. Nevertheless, neonatal brain segmentation in MRI is a challenging problem due to several factors, such as reduced tissue contrast, increased noise, motion artifacts or ongoing white matter myelination in infants.

To address this problem, a wide variety of methods have been proposed \cite{makropoulos2017review}. A popular approach uses multiple atlases to model the anatomical variability of brain tissues \cite{cardoso2013adapt,wang2014segmentation}. However, the performance of techniques based solely on atlas fusion is somewhat limited. Label propagation or adaptive methods like parametric or deformable models \cite{wang2011automatic} can be applied to refine prior estimates of tissue probability \cite{wang2011automatic}. Nevertheless, an important drawback of using such approaches for infant brain segmentation is the high risk of error due to high spatial variability in the neonatal population. Moreover, to obtain accurate segmentations, these methods typically require a large number of annotated images, which is time-consuming and requires extensive expertise.





In the last years, deep learning methods have been proposed as an efficient alternative to aforementioned approaches. Particularly, convolutional neural networks (CNNs) have been employed successfully to address various medical image segmentation problems, achieving state-of-the-art performance in a broad range of applications \cite{havaei2016hemis,kamnitsas2017efficient,DolzNeuro2017,fechter2017,wachinger2017deepnat} including infant brain tissue segmentation \cite{moeskops2016automatic,zhang2015deep,nie2016fully}. For instance, a multi-scale 2D CNN architecture is proposed in \cite{moeskops2016automatic} to obtain accurate and spatially-consistent segmentations from a single image modality.

To overcome the problem of extremely low tissue contrast between WM and GM, various works have considered multiple modalities as input to a CNN. In \cite{zhang2015deep}, MR-T1, T2 and fractional anisotropy (FA) images are merged in the input of the network. Similarly, Nie et al. \cite{nie2016fully} proposed a fully convolutional neural network (FCNN), where these image modalities are processed in three independent paths, and their corresponding features later fused for final segmentation. Yet, these approaches present some significant limitations. First, some architectures \cite{moeskops2016automatic,zhang2015deep} adopt a sliding-window strategy where regions defined by the window are processed one-by-one. This leads to a low efficiency and a non-structured prediction which reduces the segmentation accuracy. Second, these methods often employ 2D patches as input to the network, completely discarding anatomic context in directions orthogonal to the 2D plane. As shown in \cite{DolzNeuro2017}, considering 3D convolutions instead of 2D ones results in a better segmentation.


In light of above-mentioned challenges and limitations, we propose a hyper-densely connected 3D fully convolutional network, called \emph{HyperDenseNet}, for the voxel-level segmentation of infant brain in MR-T1 and T2 images. Unlike the methods presented in \cite{moeskops2016automatic,zhang2015deep,nie2016fully}, our network can incorporate 3D context and volumetric cues for effective volume prediction. The proposed HyperDenseNet network also extends our recent work in \cite{DolzNeuro2017} by exploiting dense connections in a multi-modal image scenario. This dense connectivity facilitates the learning process by including deep supervision and improving gradient flow. To the best of our knowledge, this is the first attempt to densely-connect layers across multiple independent paths, each of them specifically designed for a different image modality. We validate the proposed network on data from the iSEG-2017 MICCAI Grand Challenge on 6-month infant brain MRI Segmentation, showing the state-of-the-art performance of our network.

\section{Methodology}


\subsection{Single-path baseline}
\label{sec:baseline}

The architectures presented in this work, which are built on top of \textit{DeepMedic} \cite{kamnitsas2017efficient}, are inspired by our recent work in \cite{DolzNeuro2017}, where we proposed a 3D fully CNN to segment subcortical brain structures. An important feature of that network was its ability to model both local and global context by embedding intermediate-layer outputs in the final prediction. This helped enforce consistency between features extracted at different scales, and embed fine-grained information directly in the segmentation process. Hence, outputs from intermediate convolutional layers (i.e., layers 3 and 6) were directly connected to the first fully connected layer (\textit{fully\_conv\_1})\footnote{Fully connected layers are replaced by a set of \vold{1} convolutional filters.}. 

As baseline, we extend this semi-dense architecture to a fully-dense one, by connecting the output of all convolutional layers to \textit{fully\_conv\_1}. In this network, MR-T1 and T2 are concatenated before the input of the CNN, and processed together via a single path. Table \ref{table:layers} shows the architecture of this baseline network, where each convolutional block is composed of batch normalization, a non-linearity (PReLu), and a convolution. Due to space limitations, we refer the reader to \cite{kamnitsas2017efficient} and \cite{DolzNeuro2017} for additional details. 

\begin{table}[ht!]
\centering
\footnotesize
\renewcommand{\arraystretch}{1.1}
\begin{tabular}{lcccc}
\toprule
 & \multicolumn{1}{l}{\textbf{Conv. kernel}} & \multicolumn{1}{l}{\textbf{\# kernels}} & \multicolumn{1}{l}{\textbf{Output Size}} & \multicolumn{1}{l}{\textbf{Dropout}} \\ 
\midrule\midrule
\textbf{conv\_1}  & 3$\times$3$\times$3    & 25   & \vold{25}  & No    \\ 
\textbf{conv\_2}  & 3$\times$3$\times$3   & 25  & \vold{23}   & No  \\ 
\textbf{conv\_3}  & 3$\times$3$\times$3    & 25     & \vold{21} & No \\ 
\textbf{conv\_4}   & 3$\times$3$\times$3    & 50   & \vold{19}   & No  \\ 
\textbf{conv\_5} & 3$\times$3$\times$3    & 50     & \vold{17}   & No  \\ 
\textbf{conv\_6} & 3$\times$3$\times$3   & 50    & \vold{15}   & No   \\ 
\textbf{conv\_7} & 3$\times$3$\times$3   & 75   & \vold{13}    & No  \\ 
\textbf{conv\_8} & 3$\times$3$\times$3    & 75    & \vold{11}  & No   \\ 
\textbf{conv\_9} & 3$\times$3$\times$3    & 75   & \vold{9}  & No    \\ 
\textbf{fully\_conv\_1} & 1$\times$1$\times$1   & 400      & \vold{9}  & Yes   \\ 
\textbf{fully\_conv\_2} & 1$\times$1$\times$1   & 200    & \vold{9}   & Yes \\ 
\textbf{fully\_conv\_3} & 1$\times$1$\times$1  & 150   & \vold{9}   & Yes  \\ 
\textbf{Classification} &  1$\times$1$\times$1   & 4   & \vold{9}   & No   \\ 
\bottomrule
\end{tabular}
\caption{ Layers used in the proposed architecture and corresponding values with an input of size \vold{27}. In the case of multi-modal images, convolutional layers (conv\_x) are present in both paths of the network. All convolutional layers have a stride of one pixel.}
\label{table:layers}
\end{table}

\subsection{The proposed hyper-dense network}
\label{sec:hyperDense}


The concept of ``\emph{the deeper the better}'' is considered as a key principle in deep learning architectures \cite{he2016deep}. Nevertheless, one obstacle when dealing with deep architectures is the problem of vanishing/exploding gradients, which hamper convergence during training. To address these limitations in very deep architectures, densely connected networks were proposed in \cite{huang2017densely}. DenseNets are built on the idea that adding direct connections from any layer to all subsequent layers in a feed-forward manner makes training easier and more accurate. This is motivated by three observations. First, there is an implicit deep supervision thanks to short paths to all feature maps in the architecture. Second, direct connections between all layers help to improve the flow of information and gradients throughout the entire network. And third, dense connections have a regularizing effect, which results in a reduced risk of over-fitting on tasks with smaller training sets.
 
Inspired by the recent success of such densely-connected networks in medical image segmentation \cite{li2017h,yu2017automatic}, we propose a hyper-dense architecture, called HyperDenseNet, for the segmentation of multi-modal images. Unlike the baseline model, where dense connections are employed through all the layers in a single stream, we exploit the concept of dense connectivity in a multi-modal image setting. In this scenario, each modality is processed in an independent path, and dense connections occur not only between layers within the same path, but also between layers in different paths.

\begin{figure}[h!]
\centerline{\includegraphics[width=1\linewidth]{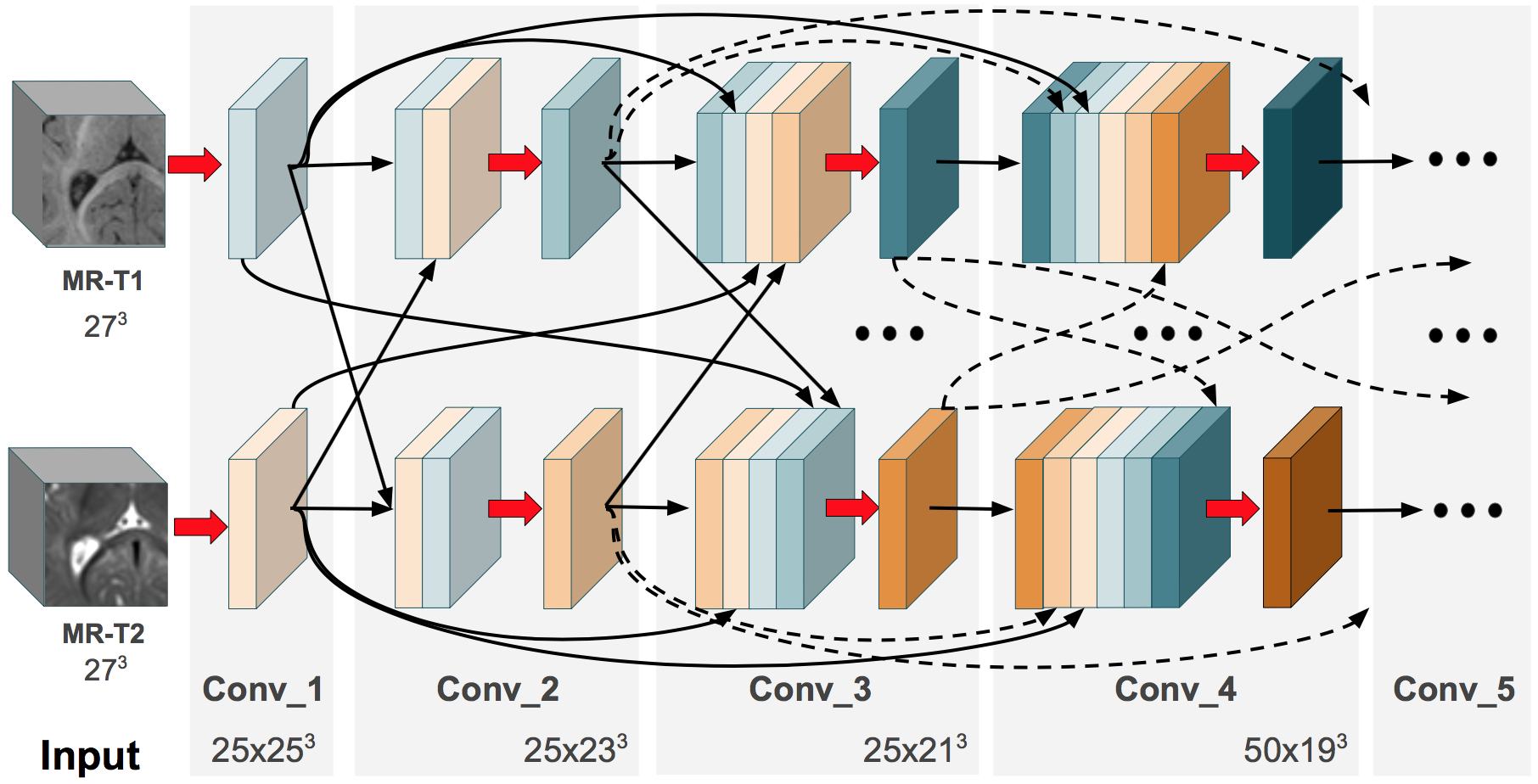}}
\caption{A section of the proposed HyperDenseNet. Each gray region represents a convolutional block. Red arrows correspond to convolutions and black arrows indicate dense connections between feature maps. Hyper-dense connections are propagated through all the layers of the network.}
\label{fig:net}
\end{figure}

The blocks composing our HyperDenseNet are similar to those in the baseline architecture. Let $\xx_l$ be the output of the $l^{th}$ layer. In CNNs, this vector is typically obtained from the output of the previous layer $\xx_{l-1}$ by a mapping $H_l$ composed of a convolution followed by a non-linear activation function:
\begin{equation}
  \xx_l \ = \ H_l(\xx_{l-1}).
 \label{eq:layer_output}
\end{equation}
In a densely-connected network, connectivity follows a pattern that iteratively concatenates all feature outputs in a feed-forward manner, i.e.
\begin{equation}
  \xx_l \ = \ H_l([\xx_{l-1}, \xx_{l-2}, \ldots, \xx_{0}]),
  \label{eq:layer_outputDense}
\end{equation}
where $[...]$ represents a concatenation operation. 

Pushing this idea further, HyperDenseNet considers a more sophisticated connectivity pattern that also links the output from layers in different streams, each one associated with a different image modality. Denote as $\xx_l^1$ and $\xx_l^2$ the outputs of the $l^{th}$ layer in streams 1 and 2, respectively. The output of the $l^{th}$ layer in a stream $s$ can then be defined as
\begin{equation}
  \xx_l^s \ = \ H_l([\xx_{l-1}^1, \xx_{l-1}^2, \xx_{l-2}^1, \xx_{l-2}^2, \ldots, \xx_{0}^1, \xx_{0}^2]).
  \label{eq:layer_outputDense}
\end{equation}


A section of the proposed architecture is depicted in Figure \ref{fig:net}, where each gray region represents a convolutional block. For simplicity, we assume that red arrows indicate convolution operations only, and that black arrows represent direct connections between feature maps from different layers. Thus, the input of each convolutional block (maps before the red arrow) consists in the concatenation of the outputs (maps after red arrow) of all preceding layers from both paths.



\subsubsection{Training parameters and implementation details}

To have a large receptive field, FCNNs typically expect full images as input. The number of parameters is then limited via pooling/unpooling layers. A problem with this approach is the loss of resolution from repeated down-sampling operations. In the proposed method, we follow the technique described in \cite{kamnitsas2017efficient,DolzNeuro2017}, where sub-volumes are used as input and pooling layers are avoided. While sub-volumes of size \vold{27} are considered training, we used \vold{35} sub-volumes during inference, as in \cite{kamnitsas2017efficient,DolzNeuro2017}. 

To initialize the weights of the network, we adopted the strategy proposed in \cite{he2015delving} that allows very deep architectures to converge rapidly. In this strategy, a zero-mean Gaussian distribution of standard deviation $\sqrt{2/n_l}$ is used to initialize the weights in layer $l$, where $n_l$ denotes the number of connections to units in that layer. Momentum was set to 0.6 and the initial learning rate to 0.001, being reduced by a factor of 2 after every 5 epochs (starting from epoch 10). Network parameters are optimized via the RMSprop optimizer, with cross-entropy as cost function. The network was trained for 30 epochs, each one composed of 20 subepochs. At each subepoch, a total of 1000 samples were randomly selected from the training images and processed in batches of size 5. 


We extended our 3D FCNN architecture proposed in \cite{DolzNeuro2017}, which is based on Theano and whose source code can be found at \href{https://www.github.com/josedolz/LiviaNET}{https://www.github.com/josedolz/LiviaNET}. Training and testing was performed on a server equipped with a NVIDIA Tesla P100 GPU with 16 GB of RAM memory. Training HyperDenseNet took around 70 min per epoch, and around 35 hours in total. Segmenting a whole 3D MR scan requires 70-80 seconds on average.


\section{Experiments and results}

\subsection{Dataset}

The dataset employed in this study is publicly available from the iSEG Grand MICCAI Challenge \cite{iSEG}. Selected scans for training and testing were acquired at the UNC-Chapel Hill and were randomly chosen from the pilot study of Baby Connectome Project (BCP)\footnote{http://babyconnectomeproject.org}. All scans were acquired on a Siemens head-only 3T scanners with a circular polarized head coil. During the scan, infants were asleep, unsedated, fitted with ear protection, and their heads were secured in a vacuum-fixation device. 

T2 images were linearly aligned to their corresponding T1 images. All images were resampled into an isotropic 1$\times$ 1$\times$1 mm$^3$ resolution. Using in-house tools, standard image pre-processing steps were then applied before manual segmentation, including skull stripping, intensity inhomogeneity correction, and removal of the cerebellum and brain stem. We used 9 subjects for training the network, one for validation and 13 subjects for testing.

\subsection{Results}

To demonstrate the benefits of the proposed HyperDenseNet, Table \ref{table:results} compares the segmentation accuracy of our architecture for CSF, GM and WM brain tissues, with that of the baseline. Three metrics are employed for evaluation: Dice Coefficient (DC), modified Hausdorff distance (MHD) and average symmetric distance (ASD). Higher DC values indicate greater overlap between automatic and manual contours, while lower MHD and ASD values indicate higher boundary similarity. 

Results in Table \ref{table:results} show HyperDenseNet to outperform the baseline. Thus, our networks yields better DC and ASD accuracy values than the baseline, for all cases. Likewise, it achieves a lower MHD for GM and WM tissues. Considering standard deviations, the accuracy of HyperDenseNet shows less variance than the baseline, again in GM and WM regions. A paired sample t-test between both configurations revealed that differences were statistically significant (p $<$ 0.05) across all the results, except for the MHD in CSF tissues (p $=$ 0.658).

\begin{table}[ht!]
\centering
\footnotesize
\renewcommand{\arraystretch}{1.1}
\begin{tabular}{lccc}
\toprule
 & DC & MHD & ASD \\
 \midrule\midrule
 & \multicolumn{3}{c}{CSF}          \\
 \cmidrule{2-4}
Baseline     & 0.953 (0.007)   & \textbf{9.296} (0.942)   & 0.128 (0.016)       \\
HyperDenseNet & \textbf{0.957 (0.007)}   & 9.421 (1.392)   & \textbf{0.119 (0.017)}        \\ \midrule
\textbf{}              & \multicolumn{3}{c}{Gray Matter}  \\ 
\cmidrule{2-4}
Baseline     & 0.916 (0.009)  & 7.131 (1.729)    & 0.346  (0.041)  \\
HyperDenseNet & \textbf{0.920 (0.008)} & \textbf{5.752 (1.078)} & \textbf{0.329 (0.041)} \\ 
\midrule
\textbf{} & \multicolumn{3}{c}{White Matter} \\
\cmidrule{2-4}
Baseline & 0.895 (0.015) & 6.903 (1.140)  & 0.406 (0.051)       \\ 
HyperDenseNet & \textbf{0.901 (0.014)}  & \textbf{6.659 (0.932)} & \textbf{0.382 (0.047)} \\ 
\bottomrule
\end{tabular}
\caption{Mean segmentation values and standard deviation provided by the iSEG Challenge organizers for the two analyzed methods. In bold is highlighted the best performance for each metric.}
\label{table:results}
\end{table}

\begin{figure}[h!]
\mbox{
        \includegraphics[width=1\linewidth]{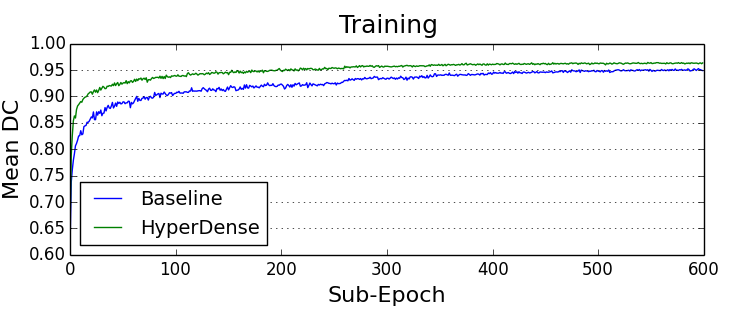}
        }
        \mbox{
        \includegraphics[width=1\linewidth]{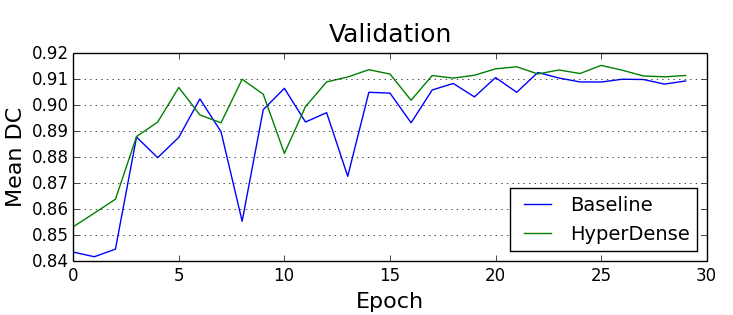}
        
        }
        \caption{ Training (\textit{top}) and validation (\textit{bottom}) accuracy plots.  }
\label{fig:training}
\end{figure}

A comparison of the training and validation accuracy between the baseline and HyperDenseNet is shown in Figure \ref{fig:training}. In these figures, mean DC for the three brain tissue is evaluated on training samples after each sub-epoch, and in the whole validation volume after each epoch. It can be observed that in both cases HyperDenseNet outperforms the baseline, achieving better results faster. This can be attributed to the higher number of direct connections between different layers, which facilitates back-propagation of the gradient to shallow layers without diminishing magnitude and thus easing the optimization.

\begin{figure}[ht!]
\mbox{ \small 
        
        \shortstack{
         \includegraphics[width=0.33\linewidth]{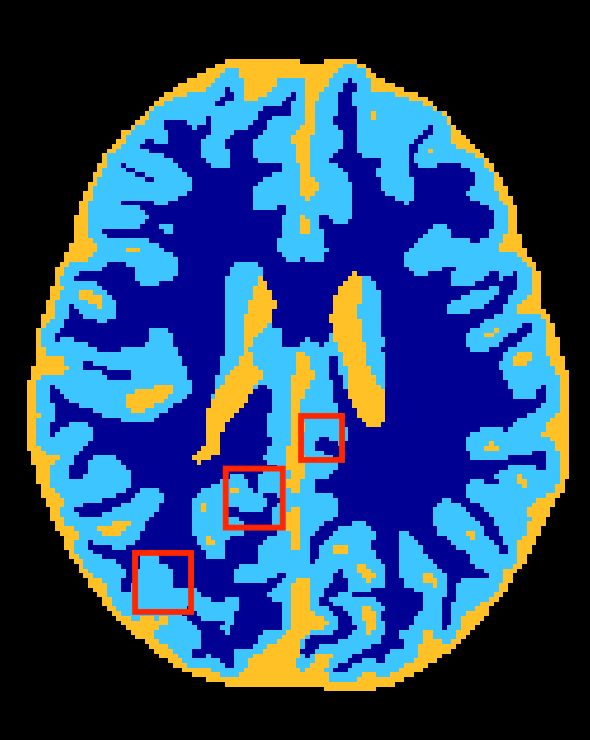} \\ Baseline
        }
        \hspace{-1.25 mm}
        \shortstack{
        \includegraphics[width=0.33\linewidth]{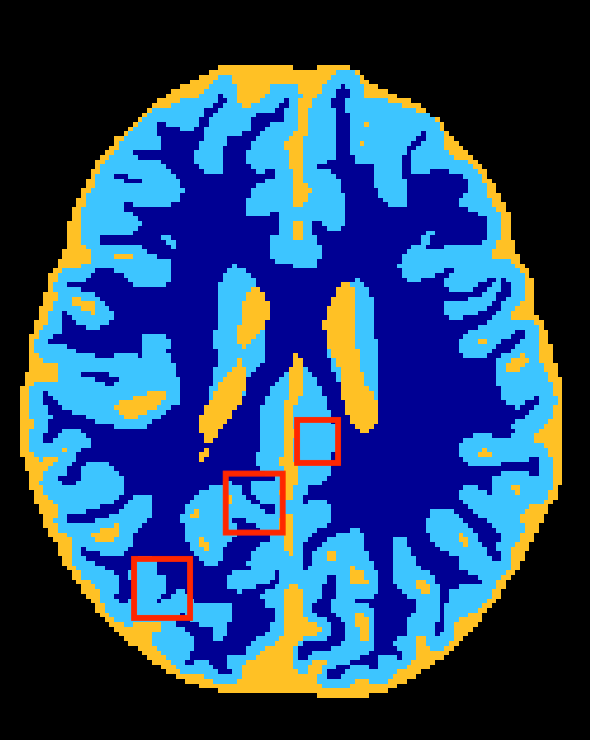} \\ HyperDenseNet
        }
        \hspace{-1.25 mm}
        \shortstack{
        \includegraphics[width=0.33\linewidth]{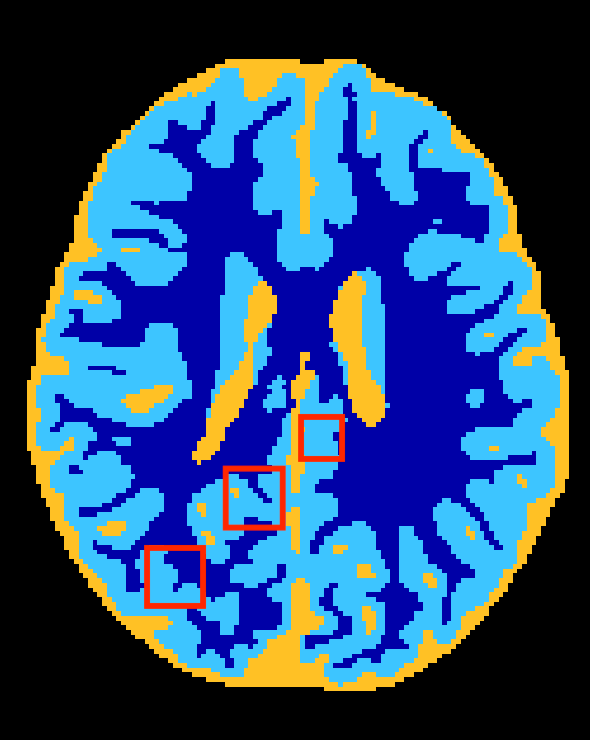} \\ Reference Contour
        }
        
              }
              
        \caption{Comparison of the segmentation results achieved by the baseline and HyperDenseNet to manual reference contour on the subject employed for validation.}
\label{fig:results}
\end{figure}

Figure \ref{fig:results} depicts visual results for the subject used in validation. It can be observed that HyperDenseNet (\textit{middle}) recovers thin regions better than the baseline (\textit{left}), which can explain improvements in distance-based metrics. As confirmed in Table \ref{table:results}, this effect is most prominent in boundaries between the gray and white matter. Further, HyperDenseNet produces fewer false positives for WM than the baseline, which tends to over-estimate the segmentation in this region.



Comparing these results with the performance of methods submitted in the first round of the iSEG Challenge \cite{iSEG}, HyperDenseNet ranked among the top-3 in 6 out of 9 metrics, being the best method in 4 of them. We can therefore say that it achieves state-of-the-art performance for the task at hand. A noteworthy point is the lower performance observed with all tested methods for the segmentation of GM and WM. This suggests that segmenting these tissues is relatively more challenging due to the unclear boundaries between them.     


An extension of this study would be to investigate deeper networks with fewer number of filters per layer, as in recently-proposed dense networks. This may reduce the number of trainable parameters, while maintaining or even improving the performance. Further, as in \cite{huang2017densely}, individual weights from dense connections could be also investigated to determine their relative importance. This would allow us to remove useless connections, making the model lighter without degrading its performance.





\section{Conclusion}


In this paper, we proposed a hyper-densely connected 3D fully CNN to segment infant brain tissue in MRI. This network, called HyperDenseNet, pushes the concept of connectivity beyond recent works, exploiting dense connections in a multi-modal image scenario. Instead of considering dense connections in a single stream, HyperDenseNet processes each modality in independent paths which are inter-connected in a dense manner.

We validated the proposed network in the iSEG-2017 MICCAI Grand Challenge on 6-month infant brain MRI Segmentation, reporting state-of-the-art results. In the future, we plan to investigate the effectiveness of HyperDenseNet in other segmentation problems that can benefit from multi-modal data.


\bibliographystyle{IEEEbib}

\bibliography{2018-ISBI}

\begin{thebibliography}{10}

\bibitem{makropoulos2017review}
Antonios Makropoulos et~al.,
\newblock ``A review on automatic fetal and neonatal brain {MRI}
  segmentation,''
\newblock {\em NeuroImage}, 2017.

\bibitem{cardoso2013adapt}
M~Jorge Cardoso et~al.,
\newblock ``{AdaPT}: an adaptive preterm segmentation algorithm for neonatal
  brain {MRI},''
\newblock {\em NeuroImage}, vol. 65, pp. 97--108, 2013.

\bibitem{wang2014segmentation}
Li~Wang et~al.,
\newblock ``Segmentation of neonatal brain {MR} images using patch-driven level
  sets,''
\newblock {\em NeuroImage}, vol. 84, pp. 141--158, 2014.

\bibitem{wang2011automatic}
Li~Wang et~al.,
\newblock ``Automatic segmentation of neonatal images using convex optimization
  and coupled level sets,''
\newblock {\em NeuroImage}, vol. 58, no. 3, pp. 805--817, 2011.

\bibitem{havaei2016hemis}
Mohammad Havaei, Nicolas Guizard, Nicolas Chapados, and Yoshua Bengio,
\newblock ``Hemis: Hetero-modal image segmentation,''
\newblock in {\em International Conference on Medical Image Computing and
  Computer-Assisted Intervention}. Springer, 2016, pp. 469--477.

\bibitem{kamnitsas2017efficient}
Konstantinos Kamnitsas et~al.,
\newblock ``Efficient multi-scale {3D} {CNN} with fully connected {CRF} for
  accurate brain lesion segmentation,''
\newblock {\em Medical image analysis}, vol. 36, pp. 61--78, 2017.

\bibitem{DolzNeuro2017}
Jose Dolz et~al.,
\newblock ``{3D} fully convolutional networks for subcortical segmentation in
  {MRI}: {A} large-scale study,''
\newblock {\em NeuroImage}, 2017.

\bibitem{fechter2017}
Tobias Fechter et~al.,
\newblock ``Esophagus segmentation in {CT} via {3D} fully convolutional neural
  network and random walk,''
\newblock {\em Medical Physics}, 2017.

\bibitem{wachinger2017deepnat}
Christian Wachinger, Martin Reuter, and Tassilo Klein,
\newblock ``Deepnat: Deep convolutional neural network for segmenting
  neuroanatomy,''
\newblock {\em NeuroImage}, 2017.

\bibitem{moeskops2016automatic}
Pim Moeskops et~al.,
\newblock ``Automatic segmentation of {MR} brain images with a convolutional
  neural network,''
\newblock {\em IEEE Transactions on Medical Imaging}, vol. 35, no. 5, pp.
  1252--1261, 2016.

\bibitem{zhang2015deep}
W.~Zhang et~al.,
\newblock ``Deep convolutional neural networks for multi-modality isointense
  infant brain image segmentation,''
\newblock {\em NeuroImage}, vol. 108, pp. 214--224, 2015.

\bibitem{nie2016fully}
Dong Nie et~al.,
\newblock ``Fully convolutional networks for multi-modality isointense infant
  brain image segmentation,''
\newblock in {\em 13th International Symposium on Biomedical Imaging (ISBI),
  2016}. IEEE, 2016, pp. 1342--1345.

\bibitem{he2016deep}
Kaiming He et~al.,
\newblock ``Deep residual learning for image recognition,''
\newblock in {\em Proceedings of the IEEE CVPR}, 2016, pp. 770--778.

\bibitem{huang2017densely}
Gao Huang et~al.,
\newblock ``Densely connected convolutional networks,''
\newblock in {\em Proceedings of the IEEE CVPR}, 2017.

\bibitem{li2017h}
Xiaomeng Li et~al.,
\newblock ``H-{D}ense{UN}et: Hybrid densely connected {UN}et for liver and
  liver tumor segmentation from {CT} volumes,''
\newblock {\em arXiv:1709.07330}, 2017.

\bibitem{yu2017automatic}
Lequan Yu et~al.,
\newblock ``Automatic {3D} cardiovascular {MR} segmentation with
  densely-connected volumetric convnets,''
\newblock in {\em International Conference on MICCAI}. Springer, 2017, pp.
  287--295.

\bibitem{he2015delving}
Kaiming He et~al.,
\newblock ``Delving deep into rectifiers: Surpassing human-level performance on
  imagenet classification,''
\newblock in {\em Proceedings of the IEEE ICCV}, 2015, pp. 1026--1034.

\bibitem{iSEG}
L.~{Wang}, D.~{Nie}, G.~{Li}, É. {Puybareau}, J.~{Dolz}, Q.~{Zhang},
  F.~{Wang}, J.~{Xia}, Z.~{Wu}, J.~{Chen}, K.~{Thung}, T.~D. {Bui}, J.~{Shin},
  G.~{Zeng}, G.~{Zheng}, V.~S. {Fonov}, A.~{Doyle}, Y.~{Xu}, P.~{Moeskops},
  J.~P.~W. {Pluim}, C.~{Desrosiers}, I.~{Ben Ayed}, G.~{Sanroma}, O.~M.
  {Benkarim}, A.~{Casamitjana}, V.~{Vilaplana}, W.~{Lin}, G.~{Li}, and
  D.~{Shen},
\newblock ``Benchmark on automatic 6-month-old infant brain segmentation
  algorithms: The iseg-2017 challenge,''
\newblock {\em IEEE Transactions on Medical Imaging}, pp. 1--1, 2019.

\end{thebibliography}

\end{document}